%% file: template.tex
\documentclass{article}

\input{math_commands.tex}

\usepackage{arxiv}

\usepackage[utf8]{inputenc} 
\usepackage[T1]{fontenc}    
\usepackage{hyperref}       
\usepackage{url}            
\usepackage{booktabs}       
\usepackage{amsfonts}       
\usepackage{nicefrac}       
\usepackage{graphicx}
\usepackage{arydshln}  

\title{A semantic hierarchical graph neural network for text classification}

\author{Shuai Hua, Xinxin Li, Yunpeng Jing, Qunfeng Liu  \\
Dongguan University of Technology, China\\
\texttt{hshuai97@qq.com}, \texttt{lixin\_2@163.com},\\ \texttt{jing.yunpeng@qq.com}, \texttt{liuqf@dgut.edu.cn}\\
}

\begin{document}
\maketitle
\begin{abstract}
The key to the text classification task is language representation and important information extraction, and there are many related studies. In recent years, the research on graph neural network (GNN) in text classification has gradually emerged and shown its advantages, but the existing models mainly focus on directly inputting words as graph nodes into the GNN models ignoring the different levels of semantic structure information in the samples. To address the issue, we propose a new hierarchical graph neural network (HieGNN) which extracts corresponding information from word-level, sentence-level and document-level respectively. Experimental results on several benchmark datasets achieve better or similar results compared to several baseline methods, which demonstrate that our model is able to obtain more useful information for classification from samples.
\end{abstract}


\section{Introduction}
Text classification, which aims to classify unknown texts into predefined categories, has always been a relatively popular research direction in natural language processing (NLP). Two very critical points in text classification are language representation and extraction of important information related to the task. Many classification methods such as CNN-based models \cite{DBLP:conf/emnlp/Kim14, DBLP:conf/nips/Zhang15char-cnn} attention-based models \cite{DBLP:conf/naacl/Yang16HAN, DBLP:conf/naacl/PetersNIGCLZ18}, transformer-based models \cite{DBLP:conf/naacl/DevlinCLT19, DBLP:conf/nips/YangDYCSL19}, have been proposed and achieved the state of the art (SOTA) results because their models represent the text input to the model well and extract enough semantic information to some extent from different semantic perspectives.

We know that from the macroscopic things as large as the cosmic celestial bodies to the microscopic things as small as molecules and atoms, we can use graphs to describe their internal relationships, and, text data is no exception. Recently, graph neural networks have shown advantages in text classification, especially news classification tasks \cite{DBLP:conf/www/PengLHLBWS018, DBLP:conf/icml/WuSZFYW19, DBLP:conf/icaart/PalSS20} due to their structural flexibility and convenient for abstract relationship between entities \cite{DBLP:conf/iclr/KipfW17}. \cite{DBLP:conf/aaai/YaoM019} modeled a dataset directly into a graph structure to obtain word-word and word-document relationships and got the best classification results at the time. However, when new samples need to be predicted, the model needs to be retrained. To solve this problem, \cite{DBLP:conf/emnlp/HuangMLZW19} built each sample in the dataset into a graph and used two global matrices to store node features and edge weight, respectively. However, these methods have two main problems: firstly, these methods ignore the hierarchical structure information of the text, that is, within the sample, the relationship between sentences or phrases is ignored; secondly, the vector representation of a word in the global matrix will not change with different context, which means that a word has only one meaning. Obviously, this is not in line with our understanding. Whether the entire dataset is directly constructed as a graph or a sample is constructed as a graph, only using nodes to represent words will lose some useful information.

According to the habit of human beings to obtain text data, for example, when people read a piece of news, they are always accustomed first to understand what each paragraph and sentence say, and then summing it up to get the final information conveyed by the news. Therefore, we should let the deep learning model learn the hierarchical representation of the data. The research on the application of graph neural networks in text classification tasks, such as the work of \cite{DBLP:conf/emnlp/HuangMLZW19}, has achieved good results, but it ignores the semantic structure information of text to a certain extent. Therefore, to address these issues, we propose a semantic hierarchical graph neural network (HieGNN) to construct graphs from word-level, sentence-level and document-level, respectively. Because of the existence of word-level, it is possible to make a word have different semantics in different sentences. As shown in Fig.1, perform word segmentation on the input text to obtain word-level graph $\mathcal{G}_1$ and sentence-level graph $\mathcal{G}_2$ and directly construct the input text into document-level $\mathcal{G}_3$ without sentence segmentation. Then use graph neural networks (GNNs) to act on the three levels mentioned above to obtain the output of the three-level vector representations, finally, merge them and input them to the softmax and linear layer for classification.
\begin{figure}[htbp]
\begin{center}
    \includegraphics[scale=0.4]{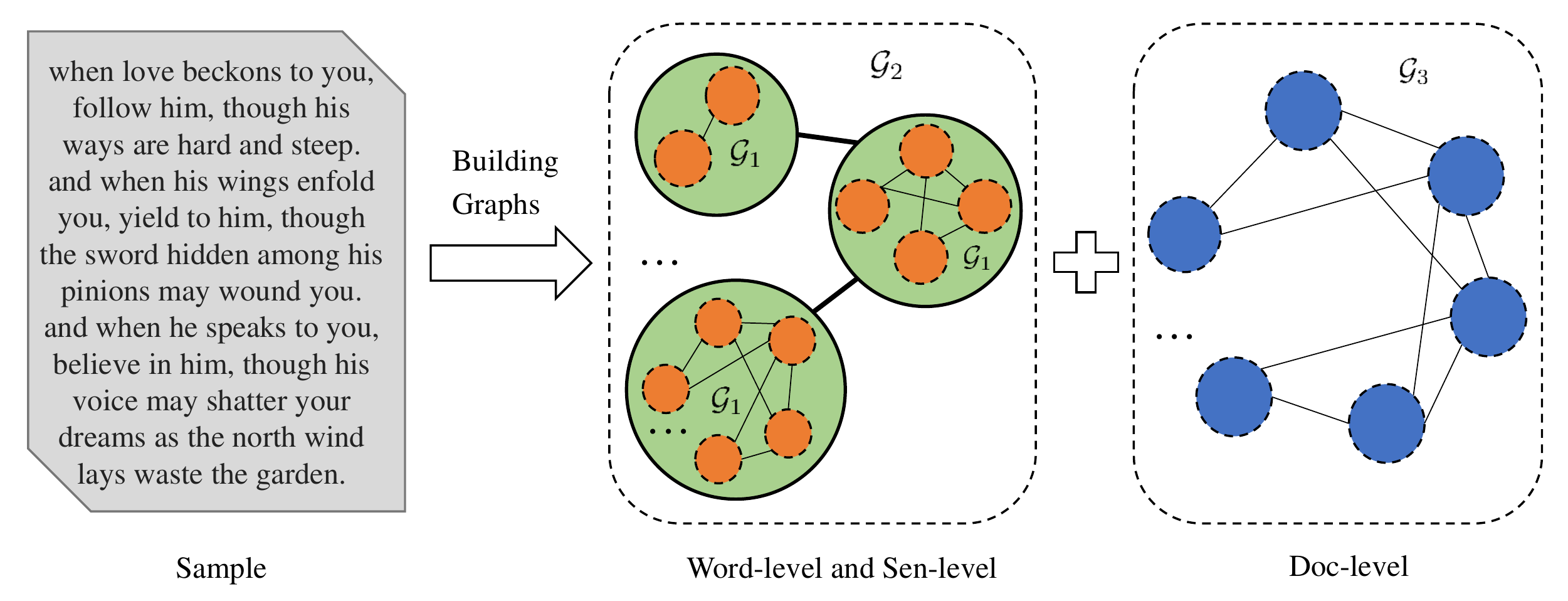}
    \end{center}
    \caption{Illustration of HieGNN processing a sample: A sample will use GNN for semantic extraction from three levels: Word-level, sen-level and doc-level, and finally the outputs of the three levels will be merged for further classification.}
\end{figure}

In this paper, we conduct experiments applying graph attention networks (GATs) \cite{DBLP:journals/corr/Velickovic17gat} at three levels mentioned above. GATs and graph convolutional networks (GCNs) are two different graph neural network models. The main difference is that GCNs use message passing mechanism (MPM) \cite{DBLP:conf/icml/Gilmer-17-MPM} when aggregating k-hop neighbor nodes, while GAT first uses the attention mechanism \cite{vaswani2017attention} to calculate the weight of the edge and then aggregates. We conducted experiments on 5 datasets, and achieved better or similar results compared with multiple baseline methods, indicating that the HieGNN model proposed in this paper can effectively extract more text information to a certain extent under the condition of using only graph neural networks. The source code will be uploaded to GitHub for reproduction and comparison. In summary, the key contributions of this paper are as follows:

\begin{itemize}
  \item In this paper, we propose a three-level hierarchical graph neural network model (HieGNN).
  \item We demonstrate the effectiveness of the proposed model HieGNN to extract textual semantic information from different semantic levels.
  \item This model is simple and not limited by specific neural networks, and can be generalized to other models and tasks.
\end{itemize}

\section{Related Work}
Text classification is one of the most important and fundamental tasks in NLP, it aims to classify new texts into predefined categories based on training samples. The study of text classification has been around for more than 60 years \cite{DBLP:journals/tist/LiPLXYSYH22}. The related research of graph neural networks (GNNs) also has a history of several decades \cite{DBLP:journals/aiopen/Zhou-20-GNN-review, DBLP:journals/tnn/Wu-21-survey-GNN}, but the research on the combination of GNNs and text classification has only emerged in recent years \cite{DBLP:conf/uemcom/Malekzadeh-21-review-GNN-TC}. The early GNNs was difficult to popularize and use due to the complexity of the algorithm. Since \cite{DBLP:conf/iclr/KipfW17} simplified the complex GNNs, the research and application of GNNs have been in full swing. And due to the flexibility of the graph, the recent performance of graph neural networks on text classification tasks is very attractive \cite{DBLP:conf/acl/LinMSHKLW21}.

Earlier, \cite{DBLP:conf/aaai/YaoM019} have successfully applied the GCNs model to the text classification task, which enables nodes to obtain k-hops neighbors' information by using a nonlinearly connected k-layer GCNs. On the basis, \cite{DBLP:conf/icml/WuSZFYW19} further demonstrated that the information acquisition of nodes in the GCNs model mainly comes from averaging the information of neighbor nodes, and removes the nonlinear transformation between layers. It significantly improves the training speed of the model and the classification effect does not drop. \cite{DBLP:conf/emnlp/HuangMLZW19} proposed a text-level GCN to solve the disadvantage that the general GNN-based models need to construct the entire corpus into a graph, which improves the classification results of GCN and reduces memory consumption to a certain extent by constructing a sample into a graph directly. However, whether it is directly constructing a sample into a graph alone, the structural information of the sample is not fully utilized. That is, words forms phrases, phrases forms sentences, sentences forms samples and even more levels can be divided. Therefore, unlike these methods, our model adopts a three-level semantic hierarchy model, which focuses on the role of the words, sentences and documents, respectively, so that the model can fully utilize the semantic structure information of the samples.

 Our work is also inspired by the hierarchical attention network (HAN) proposed by \cite{DBLP:conf/naacl/Yang16HAN}, which uses word and sentence attention mechanisms to find the most important words and sentences for the document classification tasks. The method in this paper also has certain similarities with DIFFPOOL proposed by \cite{DBLP:conf/nips/Ying18DIFFPOOL}, the key difference is that DIFFPOOL is a hierarchical pooling of a graph, while this paper decomposes the original large graph from a semantic point of view, and attempts to use three different semantic levels extraction of information. The hierarchical graph mutual information (HGMI) \cite{DBLP:journals/corr/Li-22-HGMI} and semi-supervised graph classification via cautious/active iteration (SEAL-C/AI) \cite{DBLP:conf/www/Li-19-SEAL-C-AI} analyze the advantages of hierarchical graph models from the perspective of social networks, and this paper also has similarities with them, but the details of our model are completely different. At the same time, we noticed that the GAT model proposed by \cite{DBLP:journals/corr/Velickovic17gat} shows the effectiveness of the attention mechanism on graphs, but there is a lack of related research on applying GAT to the text classification task. Therefore, in this paper, GAT is used as a specific GNN model to introduce our model.

\section{HieGNN}
In this section, we will introduce our model hierarchical graph neural network (HieGNN) in detail, which aims to utilize the semantic information of text more reasonably from different levels. In general, our model consists of three steps: constructing graphs from samples, applying GNN models, and using a linear classifier to output results.

\subsection{GAT}
In this subsection, we will use graph attention network (GAT) \cite{DBLP:journals/corr/Velickovic17gat} as an example to gradually build HieGNN without modifying the original GAT formula.  To recap, the words in a sentence in a sample are constructed as a graph at the word-level. For the sen-level, it is to construct a graph of multiple sentences in the sample. For the doc-level, each word in the sample is directly constructed into a graph. Firstly, according to the formula of GAT\cite{DBLP:journals/corr/Velickovic17gat}, we know that on a graph, the update formula of node feature is:
\begin{equation}
    \mathbf{h}_i^{(l+1)} = \sigma(\sum_{j\in Nei(i)}\alpha_{ij}^{(l)} \mathbf{z}_j^{(l)} )
\end{equation}
where, $\mathbf{h_i^{(l+1)}}$ is the updated vector representation (the $l+1$ layer of GAT) of the $i$-th node in the graph, $\sigma$ is activation function, $Nei(i)$ is the neighbor set of the $i$-th node, $\alpha_{ij}^{(l)}$ (attention score) is the weight of the edge between the $i$-th node and the neighbor node $j$, and $z_j^{(l)}$ is the vector representation of the $j$-th node in $l$ layer computed by formula (2). And the calculation formulas for $\alpha_{ij}$ are as follows:
\begin{align}
    & \mathbf{z}_i^{(l)} = \mathbf{W}^{(l)} \mathbf{h}_i^{(l)}\\
    & e_{ij}^{(l)} = LeakyReLU(\Vec{a}^{(l)^T}(\mathbf{z}_i^{(l)} || \mathbf{z}_j^{(l)}))\\
    & \alpha_{ij}^{(l)} = \frac{exp(e_{ij}^{(l)})}{\sum_{k\in Nei(i)} exp(e_{ik}^{(l)})}
\end{align}
Equation (2) is a linear transformation of the eigenvector of the $l$ layer node. "$||$" is concatenation, and $\Vec{a}^{(l)}$ is a learnable weight vector. With the above knowledge about GAT, here we apply it to the HieGNN model.

\subsection{Three Levels on GNN}
 As shown in Fig.2, firstly, we split and construct the sample into word-level graph $\mathcal{G}_1$, document-level graph $\mathcal{G}_3$ (abbreviated as doc-level), and use the word-level GNN to obtain the vector representation of a single sentence to construct sentence-level graph $\mathcal{G}_2$ (abbreviated as sen-level), and then use the sen-level GNN to obtain the semantics of sample, finally, merge the vector representations of these three levels GNN output into a final vector to represent the sample.
 
 \begin{figure}[htbp]
\begin{center}
\includegraphics[scale=0.5]{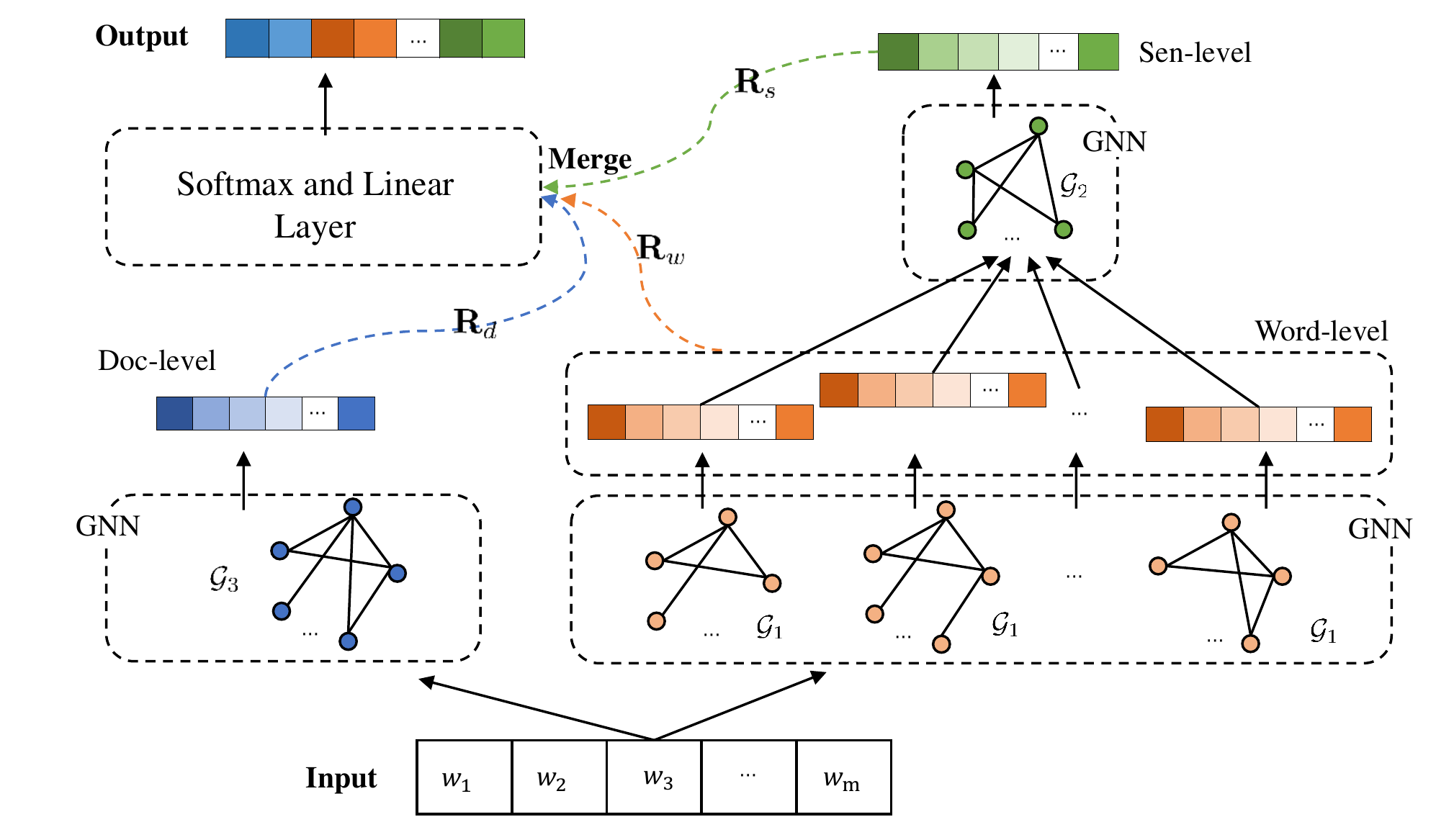}
\end{center}
\caption{An illustration of the semantics of text extracted hierarchically using GNN: In the word-level $\mathcal{G}_1$, a sentence is built into a graph, and the nodes are word tokens, and the feature of the node is the embedding vector of the word. In the sen-level $\mathcal{G}_2$, the node features are sentence vectors formed by the aggregation of $\mathcal{G}_1$ node features. In the doc-level $\mathcal{G}_3$, an example is treated as a sentence, the nodes also are word tokens, and the node features are word embedding vectors. The existence of the edge of $\mathcal{G}_1$, $\mathcal{G}_2$ and $\mathcal{G}_3$ are determined according to n-grams, and the weight of the edge is obtained according to the adaptive attention mechanism in GAT.}
\end{figure}

Suppose a sample to be classified is denoted as $X=\{w_1, w_2, \ldots, w_i, \ldots w_m\}$, $w_i$ and $m$ represent the $i$-th word and the number of words in the sample respectively. And the sample $X$ will be divided into $S= \{ s_1, s_2, \ldots, s_j, \ldots, s_k\}$ according to the punctuation in samples, $s_j$ is the $j$-th sentence and $k$ is the number of sentences in the sample. Then convert $s_j$ to a vector representation $\mathbf{H}=\{ \mathbf{h_1}, \mathbf{h_2}, \ldots, \mathbf{h_i}, \ldots, \mathbf{h_m} \}^T$ using word embeddings \cite{DBLP:journals/corr/mikolov13abs-1301-3781}, whose vector representation can be adjusted during training. $\mathbf{H} \in \mathbb{R}^{m \times n}$, $n$ is the dimension of word embedding. As shown in Fig.2: For the word-level, construct a graph $\mathcal{G}_1(V_1, E_1)$ on $s_j$ sentence, where $V_1$ represents the set of nodes, $E_1$ represents the set of edges. For the sen-level, the graph $\mathcal{G}_2(V_2, E_2)$ is constructed on $S$ and a sentence vector generated by the word-level graph $\mathcal{G}_1$ is used as node feature. For the doc-level $\mathcal{G}_3(V_3, E_3)$, the construction process is same as \cite{DBLP:conf/emnlp/HuangMLZW19}, that is, the sample $X$ is directly regarded as a sentence and constructed in a word-level manner. Below we will introduce the above processes one by one.

\textbf{Word-Level}: Observing samples from a word-level perspective, first, we build a global word embedding matrix $\mathbf{M}_1 \in \mathbb{R}^{N \times n}$ to store the graph node features based on the current dataset of this sample, where $N$ means the number of words in the vocabulary of this dataset, $n$ is the dimension of node eigenvector. We do not save edge weights because, in GAT, edge weights are adaptively determined according to the node-to-node attention scores, see equation (2)-(4). As mentioned above, then the sample $X$ is divided into multiple sentences $S=\{s_1, s_2, \ldots, s_j, \ldots, s_k \}$ based on punctuation, $s_i=\{w_1, w_2, \ldots, w_j, \ldots, w_m\}$. For every sentence, according to the $\mathbf{M}_1$ matrix, the word $w_j$ is converted into the initial feature vector $\mathbf{h_j^{(0)}}$, and the edge between nodes are determined by the n-gram (refer to Eq. (7) below). After the graph $\mathcal{G}_1(V_1, E_1)$ is constructed for every sentence, the graph node is updated according to equation (1).  The eigenvector of $s_i$ sentence is obtained according to the following formula:
\begin{equation}
    \mathbf{r}_i^{(l+1)} = R(\mathbf{h}_j^{(l)}| j\in \mathcal{G}_1)
\end{equation}
In Equation (5), $R$ is a readout function \cite{DBLP:conf/icml/Gilmer-17-MPM}, the common ones are mean, max and sum, for example, if it is mean, then $r_i^{(l+1)} = \frac{1}{m} \sum_{j=1}^{m}h_j^{(l)}$. Finally, each $r_i$ is merged by mean operation, that is, the first output about the sample at the word-level is obtained:
\begin{equation}
    \mathbf{R}_{w}^{(l+1)} = \frac{1}{n} \sum_{i=1}^n \mathbf{r}_i^{(l)}
\end{equation}
where, $R_{w}^{(l+1)}$ is the word-level output (See the orange-yellow vector in Fig. 2).

The role of word-level: As mentioned in section 1, people's habit of reading text and extracting information should first analyze each sentence and of the text in turn, then find out the key information in each sentence, and then synthesize the information found. Few people will directly check word by word, find out important word information directly from the text, and synthesize it. Therefore, it is necessary to add word-level graphs to comprehensively consider the above factors.

\textbf{Sen-Level}: Analyzing the text from sentence-level, if we treat the sample directly as a sentence like \cite{DBLP:conf/emnlp/HuangMLZW19}, then sentences with important information will be regarded as equally important. Therefore, we partition the samples into sentences $S$ (as mentioned above) and build the $\mathcal{G}_2(V_2, E_2)$ as shown in figure 2. It is no longer necessary to use a global matrix to store information on the sen-level because the initial feature vector of the $\mathcal{G}_2$ node is obtained first, and then GAT is used again on $\mathcal{G}_2$ to update the node and adjust the value of the word embedding vector in $\mathbf{M}_1$ (See the word-level paragraph above). Adjacency matrix $A\in \mathbb{R}^{k\times k}$ ($k$ is the number of sentences in a sample) of $\mathcal{G}_2$ is used to store the connection relationship of the sentence node $s_j$. It is determined by n-grams. For example, for 2 grams:
\begin{equation}
    A_{i,j}=
    \begin{cases}
    1, & |i-j| \leq 2 \\
    0, & otherwise
    \end{cases}
\end{equation}
which indicates that if the distance between two nodes is no greater than 2, they are connected. The number '1' only means that there is an edge between node $i$ and $j$, and '0' means there is no edge. The edge weight is obtained by Equation (4).

The eigenvalue of $\mathcal{G}_2$ node is the features aggregation of the nodes at the word-level is $\mathbf{r}_i$ (See Eq. (5)), and the node update formula of sen-level is:
\begin{equation}
    \mathbf{s}_{i}^{(l+1)} = \sigma (\sum_{k \in Nei(i)} \alpha_{ik}^{(l)} \mathbf{r}_k^{(l)})
\end{equation}
where, $\mathbf{s}_i^{(l+1)}$ is the updated node feature of node $i$, and $Nei(i)$ is the neighbor node of node i. For the transformation of the initial node $\mathbf{r}_i$ from $\mathcal{G}_1$ we use two new parameters $\mathbf{W}_s$ and $\Vec{b}$, because the model learns different information at sen-level and word-level. Here is the transform:
\begin{align}
    & \mathbf{r}_i^{(l)} = \mathbf{W}_s^{(l)} \mathbf{r}_i^{(l)}\\
    & e_{ij}^{(l)} = LeakyReLU(\Vec{b}^{(l)^T}(\mathbf{z}_i^{(l)} || \mathbf{z}_j^{(l)}))
\end{align}
Finally, merge the graph nodes feature to obtain the second output of the sample at sen-level:
\begin{equation}
    \mathbf{R}_s^{(l+1)}= \frac{1}{n} \sum_{i=1}^n \mathbf{s}_i^{(l)}
\end{equation}
The role of the sen-level solves the disadvantage that a word has only one vector representation from beginning to end in a classification task \cite{DBLP:conf/aaai/YaoM019, DBLP:conf/emnlp/HuangMLZW19}, and at the same time obtains different semantic level information of the text.

\textbf{Doc-Level}: Observing the sample from the doc-level, no matter how many sentences there are in the sample, it is treated as a long sentence, that is, $X=\{w_1, w_2, \ldots, w_i, \ldots, w_m\}$, which is also the practice of \cite{DBLP:conf/emnlp/HuangMLZW19}. In the HieGNN model, we keep this part because this layer is very useful in extracting textual information, this can be verified in the experimental section below.

For the construction of graph $\mathcal{G}_3(V_3, G_3)$, it is roughly the same as the construction of $\mathcal{G}_1$ in word-level, and also constructs a global matrix $\mathbf{M}_2 \in \mathbb{R}^{N \times n}$, and according to the n-gram to determine the node connection relationship. The application of GAT is the same as the word-level.

Finally, the doc-level output extracted from the sample is obtained by:
\begin{equation}
    \mathbf{R}_d^{(l+1)}= \frac{1}{m} \sum_{i=1}^m \mathbf{h}_i^{(l)}
\end{equation}

\subsection{Model Outputs}
As shown in Fig. 2, first convert the $n$-dimension of $\mathbf{R}_t$ to $C$-dimension using a linear layer to get the score for each category $x_i$ ($t\in \{d, s, w\}$, d, s and w represent the output of doc-level, sen-level and word-level respectively. $C$ is the number of categories in the dataset). Then the $\mathbf{R}_t \in \mathbb{R}^{1 \times C}$ is further transformed by the softmax and log function $\log\softmax(x_i) = \ln (\frac{exp(x_i)}{\sum_j^{C} exp(x_j)})$ to obtain $\mathbf{R}^{'}_t \in \mathbb{R}^{1 \times C}$. Then the outputs on the three levels are combined:
\begin{equation}
    \mathbf{\hat{y}} = \lambda_d \mathbf{R}_d^{'} + \lambda_s \mathbf{R}_s^{'} + \lambda_w \mathbf{R}_w^{'}
\end{equation}
where, $\mathbf{\hat{y}} \in \mathbb{R}^{1 \times C}$, $\lambda_t$ represents the weight of the output at the corresponding level, and $\lambda_d + \lambda_s + \lambda_w = 1$ ($\lambda_t \in (0, 1), t\in \{d, s, w\}$).

In order not to increase the complexity of the model, we only associate $\lambda$ with the number of sentences (denoted as $x_s$) in the sample, $\lambda_s$ and $\lambda_w$ decrease as the number of sentences increases, while $\lambda_d$ does the opposite:
\begin{align}
    & \lambda_d = \frac{1}{\ln(x_s) + 1} \\
    & \lambda_s = \frac{2}{3} (1-\lambda_d) \\
    & \lambda_w = \frac{1}{3} (1-\lambda_d)
\end{align}
The design motivation of Eq. (15)-(17): give less weight to word-level and sen-level with large number of sentences, this is because we prefer the model to extract the structural information in each sentence, and pay less attention to the words themselves, on the contrary, giving more weight to doc-level makes it possible to directly extract important word information from the samples.

\subsection{Optimization Goal}
After obtaining the output $\hat{\mathbf{y}}$, we use the cross-entropy loss function to optimize the model:
\begin{equation}
    \mathcal{L}=-\frac{1}{N} \sum_i^N \sum_k^{C} \mathbf{y}_{ik} \ln(\hat{\mathbf{y}}_{ik})
\end{equation}
In Eq. (18), $N$ is the number of samples in the mini-batch, and $C$ is the number of categories, and $\mathbf{y}$ are the true labels, and $\mathbf{\hat{y}}$ are the labels predicted by the model. Adjust the matrix $\mathbf{M}_1, \mathbf{M}_2$ (as mentioned above) and the parameters of the GAT model at each level according to the minimum loss $\mathcal{L}$, see Eq. (1)-(4).

\subsection{Time Complexity}
Since the HieGAT proposed in this paper reuses GAT model many times, it is necessary to analyze their time complexity. The core of the HieGAT model is GAT, which can be regarded as a repeated use of GAT to a certain extent. The main operation of GAT is Eq. (2)-(4). For one sample, assuming that $H$ is used to represent the dimension of the graph node vector, and the dimension of node vector does not change after linear transformation. So the complexity of Eq. (2) is $\mathcal{O}(H^{2})$, since each node of graph needs to be calculated, it is $\mathcal{O}(|V|\times H^{2})$ ($|V|$ is the number of graph nodes). Eq. (3) is a mapping function that maps a $2H$-dimensional vector to a real number, so its complexity is $\mathcal{O}(H)$. Eq. (4) needs to calculate the attention score for each edge of the graph, so the complexity is $\mathcal{O}(|E|\times H)$ ($|E|$ is the number of edges of graph). To sum up, the complexity of the GAT model is $T(|V|, |E|, H)=\mathcal{O}(|V|\times H^{2} +|E| \times H)$. 

For the three-level graph on HieGAT proposed in this paper, assuming that a sample is divided into $|S|$ sentences, each sentence has $|W|$ words, then for the complexity of the doc-level GAT is $T_d(|S|, |W|, H) = \mathcal{O}((|S|\times |W|\times H^2 + |E_d|\times H))$. Similarly, we get sen-level complexity $T_s(|S|, |W|, H)=\mathcal{O}(|S|\times H^2+ |E_s|\times H)$ and word-level complexity $T_w(|S|, |W|, H)= \mathcal{O}(|S|\times|W|\times H^2+|E_w|\times H)$. Since the average number of sentences per sample $|S|$ is very small (see Table 1) and n-gram are used to build the graph (See Eq. (7)), the number of edges in the graph is much smaller than $|V|\times (|V|-1)$ (according to the relationship between the number of vertices and edges of the graph). Assuming that $|E|\propto|V|$, then $|E_d|>|E_w|$, so $T_d>T_w \gg T_s$. And because at the doc-level and word-level are two computations that can be performed at the same time, the total time complexity is a little bit larger than $T_d$.

\section{Experiments}
In the previous sections, we introduced the implementation details of HieGNN using GAT as a specific model. In this section, we will use HieGAT, HieGCN, and BERTHieGAT to conduct experiments on 5 common datasets and compare them with GAT, GCN, BERT and several currently state-of-the-art models to test the effectiveness of the three-level semantic extraction method (HieGNN) proposed in this paper.

\subsection{Baseline Methods}
These baseline models we selected are the current models that apply GNN in the field of text classification, mainly TextGCN, SGC, Text-Level-GCN, BERTGCN, RoBERTaGCN, BERTGAT and RoBERTaGAT, etc. It should be pointed out that, except the Text-Level-GCN is a graph level classification model, others are graph node level classification models, and the HieGAT used in this paper is a graph level model too. 
\begin{itemize}
  \item \textbf{TextGCN }\cite{DBLP:conf/aaai/YaoM019}: The model mainly completes two tasks: the first is graph node classification and the second is graph representation learning. By constructing a dataset into a graph, the vector representation of each document node is learned. It became one of the best graph neural network models for classification at that time.
  \item \textbf{SGC} \cite{DBLP:conf/icml/WuSZFYW19}: This model mainly simplifies the current GCN model (such as FastGCN \cite{DBLP:conf/iclr/ChenMX18}), which accelerates the model inference speed without reducing the model classification effect. Even better results are achieved on some datasets compared with TextGCN.
  \item \textbf{Text-Level-GCN} \cite{DBLP:conf/emnlp/HuangMLZW19}: This model mainly models a single sample into a graph, saves global information through word embedding matrix and edge weight matrix, and finally mergers graph nodes into a vector to represent the sample, achieves a good result.
\end{itemize}

\subsection{Datasets}
In order to compare the performance of GNN and HieGNN on text classification tasks, we choose the same datasets as \cite{DBLP:conf/aaai/YaoM019, DBLP:conf/emnlp/HuangMLZW19, DBLP:conf/acl/LinMSHKLW21, DBLP:conf/icml/WuSZFYW19}. The datasets can be found here\footnote{\url{https://github.com/yao8839836/text_gcn/tree/master/data}} and the overall datasets statistics information as shown in Table 1, it should be pointed out that for the R8 and R52 datasets, the original data with punctuation has been lost, and only the cleaned data with punctuation removed is retrained in the dataset used now. This paper uses the NNSplit \footnote{\url{https://bminixhofer.github.io/nnsplit/}} library for sentence segmentation of unpunctuated text. The 20NG, Ohsumed, and MR datasets use NLTK \footnote{\url{https://www.nltk.org/}} for sentence segmentation, and all datasets use NLTK for word segmentation. For more detailed information about these datasets, refer to \cite{DBLP:conf/aaai/YaoM019}.
\begin{table}[htbp] 
  \centering
  \caption{Datasets statistics. Except the last column is newly added, other columns are taken from \cite{DBLP:conf/aaai/YaoM019}. '\#' means quantity, "Avg. L" and "Avg. S" denote the average length and the average number of sentences of each sample, respectively.}
  \begin{tabular}{cccccccc}
    \hline
    Dataset & \#Docs & \#Training & \#Test & \#Words & \#C & Avg. L & Avg. S \\
    \hline
    20NG & 18,846 & 11,314 & 7,532 & 42,757 & 20 & 221.26 & 4.89 \\
    R8 & 7,674 & 5,485 & 2,189 & 7,688 & 8 & 65.72 & 6.24 \\
    R52 & 9,100 & 6,532 & 2,568 & 8,892 & 52 & 69.82 & 6.29 \\
    Ohsumed & 7,400 & 3,357 & 4,043 & 14,157 & 23 & 135.82 & 9.02 \\
    MR & 10,662 & 7,108 & 3,554 & 18,764 & 2 & 20.39 & 1.19 \\
    \hline
  \end{tabular}
\end{table}

\subsection{Experimental Setup}
 \textbf{Experimental environment}: The experiments are mainly done using the Deep Graph Library (DGL) \footnote{\url{https://www.dgl.ai/}} in pytorch (as mentioned above), and all experiments are done on the Google Colab platform using a Tesla P100 GPU with memory 16GB.
 
 \textbf{Parameter settings}:  Use a 3-layer and 3-head GAT on the doc-level, and a 1-layer and 1-head GAT on the sen-level and word-level. The batch size is 64 and should not be set too large. (Although large batches help to speed up training, large batch size will bring a problem, that is, in a batch, $x_s$ is obtained by averaging the samples in the batch, which will affect the value of $\lambda_t$, see Eq. (15)-(17).). The activation function between GAT layers uses ELU, the negative slope of LeakyReLU angle is 0.2, and dropout rate is 0.5. The learning rate on MR dataset is 0.0001, the others are 0.001.

\textbf{Evaluation Metric}: In order to compare with related work, the accurate value is used as the evaluation metric in this paper:
\begin{equation*}
    Accuracy=\frac{\# Correct}{\# Total}
\end{equation*}
which means the number of correct classifications divided by the total number of samples.

\subsection{Experimental Results}
The performance of the benchmark methods in the above section on these 5 datasets is summarized and compared as shown in Table 2. It can be seen from the experimental results that the effect of the HieGAT model has a certain improvement on R8 and MR datasets compared with GNN-based models. And outperforms TextGCN and SGC on four out of five datasets.
\begin{table}[htbp]
    \centering
    \caption{Experimental results on 5 datasets, accuracy metric is adopted. We run 5 times and report mean results. Except for the bottom data, other data are taken from the results in the baseline papers. '-' indicates that the original paper did not provide the result.}
    \begin{tabular}{cccccc}
        \hline
        Model & 20NG & R8 & R52 & Ohsumed & MR \\
        \hline
        TextGCN & 0.8634 & 0.9707 & 0.9356 & 0.6836 & 0.7674 \\
        SGC & 0.885 & 0.972 & 0.94  & 0.685 & 0.759 \\
        TextLevelGCN & - & 0.978 & 0.946 & 0.6994 & - \\
        \hdashline
        HieGAT & 0.8584 & \textbf{0.9783}  & 0.9454 & 0.6984 & \textbf{0.7804}\\
        \hline
    \end{tabular}
\end{table}

On our proposed model, in addition to the experimental results above compared with other models, we also did comparative experiments to observe whether the hierarchical graph neural network works. Compared with HieGAT, only the value of $\lambda_t$ $(t\in \{d, s, w\})$ is changed, and other parameters remain unchanged. The main comparative experiments on 5 datasets are:
\begin{itemize}
    \item Use only one of the three levels at a time (i.e. set one coefficient to 1 and other two to 0).
    \item Use only two of the three levels at a time (i.e. set one coefficient to 0 and other two to non-zero).
\end{itemize}
The results of the ablation experiments are shown in Table 3, in which the top column is the use of only one level, the middle column is the use of only two levels, and the bottom column is the result of using three levels. When $\lambda_d=1$ and $\lambda_{s, w}=0$, it represents the classification result of the original GAT model. We can see that it is difficult to achieve the effect of HieGAT whether only one level of GAT is used alone or two levels of GAT are used. Therefore, processing the input text from a three-level perspective proposed in this paper can extract more information, thereby improving the classification performance of the original GAT model.

\begin{table}[htbp]
    \centering
    \caption{Ablation experiments on HieGAT.}
    \begin{tabular}{cccccc}
        \hline
         $\lambda$ & 20NG & R8 & R52 & Ohsumed & MR  \\
         \hline
         $\lambda_d = 1, \lambda_{s, w}=0$ & 0.8420 & 0.974 & 0.9364 & 0.6634 & 0.7731  \\
         $\lambda_s = 1, \lambda_{d, w}=0$ & 0.8064 & 0.9653 & 0.9124 & 0.6667 & 0.7597  \\
         $\lambda_w = 1, \lambda_{d, s}=0$ & 0.8033 & 0.9691 & 0.9294 & 0.6696 & 0.7640  \\
         \hdashline
         $\lambda_d=0, \lambda_{s, w} \neq 0 $ & 0.8062 & 0.9679 & 0.9204 & 0.6714 & 0.7586  \\
         $\lambda_s=0, \lambda_{d, w} \neq 0$ & 0.8551 & 0.9751 & 0.9441 & 0.6882 & 0.7659  \\
         $\lambda_w=0, \lambda_{d, s} \neq 0$ & 0.8553 & 0.9760 & 0.9392 & 0.6863 & 0.7690  \\
         \hdashline
         HieGAT$(\lambda_{d, s, w}\neq 0)$ & \textbf{0.8584} & \textbf{0.9783}  & \textbf{0.9454} & \textbf{0.6984} & \textbf{0.7804}\\
         \hline
    \end{tabular}
\end{table}

\section{Conclusion}
This paper proposes a hierarchical graph neural network (HieGAT) to solve the problem of insufficient information extraction (ie, whether a corpus is directly constructed as a graph or a sample is directly constructed as a graph, it ignores the progressive relationship of words that form sentences and sentences that form samples, and we can also further divide them into more levels) of input text by current graph neural networks (GNNs) models. By constructing a sample into three levels of corresponding graphs, the progressive semantic information of the input sample (ie, the semantics of words, the semantics of sentences, and the semantics of documents) is effectively extracted. At the same time, since our model is not related to specific GNNs, it can be easily combined with other models.

Experimental results show that our model has significant improvement on 4/5 datasets compared to the traditional GNN model TextGCN and SGC. And the computational complexity analysis of the HieGAT is slightly larger than the original GAT model, but this can be solved in the parallel operation mechanism.
Although the hierarchical structure effectively extracts textual semantics, the effect of improvement is very limited, which shows that the simple method of constructing graphs by segmenting sentences needs to be further improved.

\bibliographystyle{unsrt}  
\bibliography{references}  

\end{document}

%% file: math_commands.tex
\usepackage{amsmath,amsfonts,bm}

\def\eqref#1{equation~\ref{#1}}

\def\1{\bm{1}}

\DeclareMathAlphabet{\mathsfit}{\encodingdefault}{\sfdefault}{m}{sl}
\SetMathAlphabet{\mathsfit}{bold}{\encodingdefault}{\sfdefault}{bx}{n}

\newcommand{\softmax}{\mathrm{softmax}}

